\documentclass{article}

\usepackage{arxiv}
\usepackage[breaklinks,colorlinks]{hyperref}
\usepackage[utf8]{inputenc} % allow utf-8 input
\usepackage[T1]{fontenc}    % use 8-bit T1 fonts
\usepackage{hyperref}       % hyperlinks
\usepackage{url}            % simple URL typesetting
\usepackage{booktabs}       % professional-quality tables
\usepackage{amsfonts}       % blackboard math symbols
\usepackage{nicefrac}       % compact symbols for 1/2, etc.
\usepackage{microtype}      % microtypography
\usepackage{lipsum}
\usepackage{graphicx}
\graphicspath{ {./} }

\title{A Brief Overview of Physics-inspired Metaheuristic Optimization Techniques}

\author{
 Soumitri Chattopadhyay \\
  Department of Information Technology\\
  Jadavpur University\\
  Kolkata, India \\
   \And
 Aritra Marik \\
  Department of Information Technology\\
  Jadavpur University\\
  Kolkata, India \\
  \And
 Rishav Pramanik \\
  Department of Computer Science and Engineering\\
  Jadavpur University\\
  Kolkata, India \\
}

\begin{document}
\maketitle
\begin{abstract}
Metaheuristic algorithms are methods devised to efficiently solve computationally challenging optimization problems. Researchers have taken inspiration from various natural and physical processes alike to formulate meta-heuristics that have successfully provided near-optimal or optimal solutions to several engineering tasks. This chapter focuses on meta-heuristic algorithms modelled upon non-linear physical phenomena having a concrete optimization paradigm, having shown formidable exploration and exploitation abilities for such optimization problems. Specifically, this chapter focuses on several popular physics-based metaheuristics as well as describing the underlying unique physical processes associated with each algorithm. 
\end{abstract}

% keywords can be removed
\section{Introduction}
\label{Intro}
% Computationally challenging optimization-based problems have always been of special interest by various researchers around the globe. Such challenges differ in their types, therefore there is no ultimate best optimizer for every other challenge. A trade-off between exploration and exploitation ensures that these algorithms work well specifically for the problems it was designed for. We have observed a lot of influxes of algorithms in the recent past to fulfil the frequent requirements of optimized solutions in different domains field of Artificial Intelligence (AI)-based systems. The vastness in the diversity of these algorithms has been a reason behind researchers classifying them based on their inspirations. This has helped them with a more organised approach towards studying the algorithms and drawing implications from them. \textcolor{red}{Such is the pool of such algorithms that, recently grouping the algorithms based on their inspirations has helped researchers to organise and study its implications properly.}\\

Computationally challenging optimization problems have always been of special interest by various researchers around the globe. This is primarily due to them often having a very high dimensional search space, or having highly complex and non-linear objective functions at their core, which classical gradient-based methods fail to tackle efficiently. This has been the main reason for the development of metaheuristic algorithms that take inspiration from our surroundings (nature, swarms and physical processes) and can provide a computationally cheap yet robust optimization procedure for such hard problems at hand. Parallely, researchers have also noticed the undeniable success of modeling physics' processes to study highly complex phenomena, both in real-world and computer science. For instance, resource allocation problem has been well tackled by statistical mechanics models \cite{chakraborti2015statistical}, while certain aspects of statistical thermodynamics have been employed to explain micro-evolution of species \cite{demetrius2013boltzmann}, and so on. As a result, in spite of swarm-inspired algorithms \cite{kennedy1995particle, mirjalili2014grey} being in the forefront as robust optimizers, researchers have shown keen interest in adapting principles and theories of physics and applying them to solve real-world optimization problems.\\
Recently, the world of metaheuristics has seen the advent of several novel search mechanisms based on various non-linear physics processes. The novelty of these approaches lies in the fact that the non-linear physical phenomena are leveraged as backbones to be modeled upon in order to formulate efficient search algorithms, whose mechanism is quite different from the conventional swarm and evolutionary algorithms. Such physics-inspired algorithms have shown great promise and robustness as global optimization strategies. \\
In this chapter, we aim to discuss some of the most popular optimization algorithms derived from non-linear physics. As such, physics has several categorical subdomains of study, such as classical mechanics \cite{formato2007central, rashedi2009gsa}, thermodynamics \cite{moein2014kgmo, kaveh2017novel} and optics \cite{kaveh2012new}, to name a few. Accordingly, we have grouped the existing metaheuristics into such classes based on the physical phenomena they were modeled upon. In our discussion, we have explained the intuitions behind the algorithms, drawing parallels between the algorithmic steps and the original physical processes. Our discussion includes widely-used as well as recently proposed algorithms, so as to give the readers a holistic overview regarding physics-based metaheuristics. Specifically, we cover 21 algorithms across 6 sub-domains of physics in the present chapter.

% The inspiration for such optimization algorithms are very often inspired by observing the behaviour exhibited by any object in nature, such as ants, grey wolves, genetic reproduction, to name a few. One such field is the study of various physical phenomena around us or physics. Specifically, non-linear physics or in layperson’s terms, the physical processes which exhibit non-linear behaviour to perform a particular task have been a source of great inspiration for researchers to develop state-of-the-art algorithms for computationally challenging optimization problems.\\
% This chapter specifically attempts to focus on non-linear physics-inspired optimization algorithms, developed in the past to solve computationally challenging optimization problems. We group each of the algorithms according to the physical phenomenon associated with it and detail the inspiration behind each algorithm along with a brief discussion of the steps involved in obtaining the optimum value. \textcolor{red}{ We detail the inspirations in the present chapter, along with we discuss the steps involved in obtaining the optimum value in brief.} Specifically, we cover 21 algorithms of 6 categories in the present chapter, we group the algorithms for the convenience of the readers.

\section{Classical Mechanics based Metaheuristics}
\label{Classical}
The following section presents several optimization techniques based on classical mechanics, gravitation and kinematics. They are Central Force optimization,Gravitational search algorithm, Colliding bodies algorithm and Equilibrium optimizer.

\subsection{Central Force Optimization}
\label{CFO}
Formato et al. \cite{formato2007central} introduced Central Force Optimizer (CFO), a deterministic algorithm based on gravitational kinematics. For a deterministic, we provide the positive and negative curvature direction, whereas, for a stochastic based optimizer, we do not give these conditions in advance.\\
The CFO algorithm uses probes as its population. These probes with iterations move closer to the fittest probes, searching the search space effectively. CFO defines each probe as having a position vector, an acceleration vector, and a fitness value. The position vector is a representation of the probe’s current coordinates regarding each dimension of the search space. Under minimum storage requirements, both the previous and current positions should be stored.\\
CFO is an eight-step algorithm: the steps include initialization of position and acceleration, calculating initial probe position, calculating initial fitness values, updating probe positions, retrieving errant probes, calculating fitness values of the probes, computing new accelerations and finally like any other algorithm checking if the stopping criteria are met. The performance of the algorithm largely matters on the initial probe distribution.

\subsection{Gravitational Search Algorithm}
\label{GSA}
Inspired by the dynamics of gravity and mass interaction, Rashedi et al. \cite{rashedi2009gsa} proposed the Gravitational Search Algorithm (GSA), which is formulated on the laws of gravity and Newton’s second law of motion. In GSA, each of the particles accelerates its velocity inversely proportional to $R$ which is contrary to the law of gravitation, which says acceleration is inversely proportional to $R^2$. The authors found experimentally that use of either yields better results.

\begin{figure}[ht!]
    \centering
    \includegraphics[keepaspectratio]{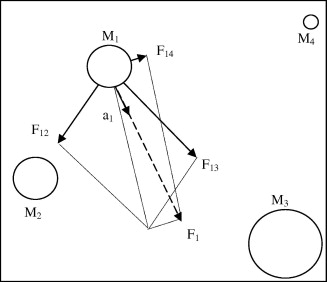}
    \caption{The influence of particle on other particles. \textit{Reproduced with permission. Original figure by \cite{rashedi2009gsa}}}
    \label{fig:GSA}
\end{figure}

\noindent The concept and formulation of GSA is similar to CFO, except that GSA is stochastic unlike the deterministic latter. The particles (masses) are influenced by the position of all the other masses. Masses have four specifications: position, inertial, active, and passive gravitational mass. The position of the mass corresponds to a solution to the problem, and its gravitational and inertial masses are determined using a pre-defined fitness function. The influence of particles can be seen in Fig \ref{fig:GSA}.\\
The process of GSA comprises eight principal steps of initialisation, fitness evaluation, calculation of gravitational constant, updating inertial and gravitational masses, updating total force, computing the acceleration and velocities, updating particle positions and finally checking if the termination criteria are achieved or not.

\subsection{Colliding Bodies Optimization}
\label{CBO}
Kaveh et al. \cite{kaveh2014colliding} proposed Colliding bodies optimization (CBO), which is inspired by the phenomenon of collision between two bodies in one dimension. Unlike other algorithms, CBO does not use any memory or any internal parameter to save its best-obtained solutions. One object collides with another object to attain minimum energy. The authors originally proposed this algorithm to solve engineering structure-based problems.\\
We refer each agent in CBO to as colliding bodies. Each of these colliding bodies has a mass and a velocity. CBO essentially comprises six principal steps of initialization, calculating the mass of each of the colliding bodies, then the entire population is sorted and is divided into two equal groups. One group starts from best and the other group starts from the middle. We consider the first group being stationary and the other group to be mobile and move towards the best-obtained solutions. Then, the new velocities are calculated after the collisions are done. Then the final positions are updated and finally, the termination criteria are checked.

\subsection{Equilibrium Optimizer}
\label{EO}
Equilibrium Optimizer (EO) was proposed by Faramarzi et al. \cite{faramarzi2020equilibrium}, the main inspiration of EO is dynamic mass balance in physics. The authors used mass balance equation primarily to mathematically model the mass balance phenomenon. The search agents are consist of concentrations which are used to find the optimal solution in the search space.\\
Just like any other metaheuristic optimization algorithm, in order to explore the maximum search space we randomly initialise the search space. An equilibrium pool is created out of the four best performing solutions and the average of the these best four solutions. Next to update the concentrations a balance between exploration and exploitation is maintained by setting parameter which is a function of number of iterations. This is done to maximise the local search ability in the later stage and enhance the exploration during the initial stages. The generation rate is set, following the first order decay process to enhance to exploitable nature of the solution. Finally the stopping criteria is checked.\\
\section{Fluid Mechanics based Metaheuristics}
\label{Fluid}
There have been some interesting optimization approaches based on the theories of fluid mechanics. The ones which have been discussed in this section are: Vortex search algorithm, Flow regime algorithm, and Archimedes optimization algorithm.

\subsection{Vortex Search Algorithm}
\label{VS}
\begin{figure}[!ht]
    \centering
    \includegraphics[scale=0.8]{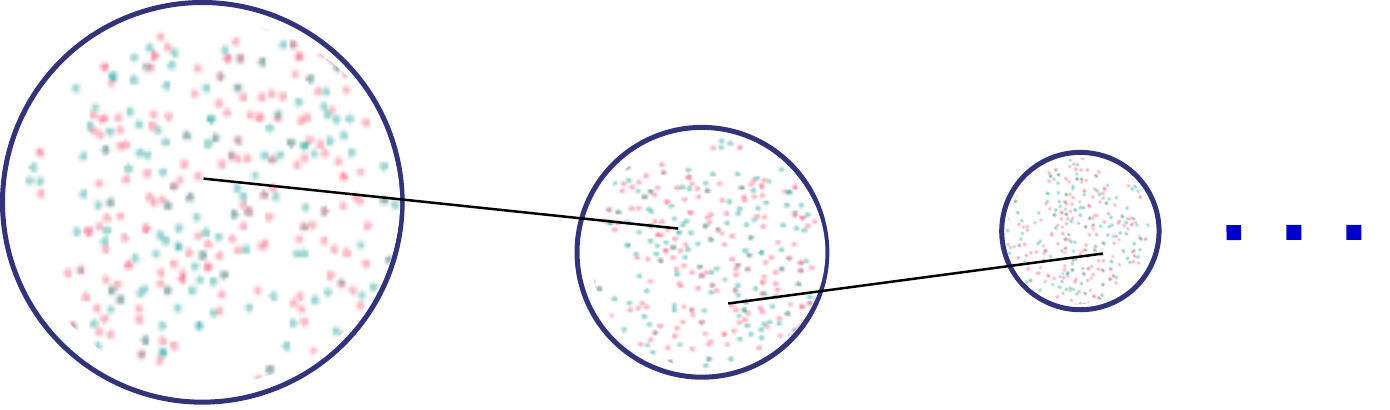}
    \caption{Intuitive behaviour of the vortex search algorithm.}
    \label{fig:vortex}
\end{figure}

Vortex search (VS) algorithm was proposed by~\cite{dougan2015new}, taking inspiration from the vortex-like phenomena occurring in irrotational, incompressible fluids. The algorithm starts with a random central point as the starting solution and generates a circular bounded (3D sphere/hypersphere for higher dimensions) Gaussian distribution around its coordinates. The candidate solutions are randomly generated from the distribution neighbourhood of the current circle center. Out of this local pool, the best candidate solution is chosen as the center for the distribution to be generated in the next step. With each iteration, the radius of the spherical distribution to be created is decreased using the inverse of the incomplete gamma function. This radius decrement operation is the key to changing the focus of the search mechanism. Thus, the intuition of the algorithm is to find vortex-like search patterns that gradually converge to an optimal point, which has been depicted in Fig. \ref{fig:vortex}.

\noindent It is important to note that contrary to other physics-based metaheuristics which are mostly population-guided, VS is a single solution-based search procedure. As a result, although the algorithm seeks to find the global optimum, the search mechanism essentially follows a local exploitation paradigm for the same.

\subsection{Flow Regime Algorithm}
\label{FRA}
Proposed in \cite{tahani2019flow}, flow regime algorithm (FRA) derives from principles of hydrodynamics for its optimization metaphor. Specifically, it models the turbulent and laminar flow regimes of a viscous fluid to define its exploration and exploitation mechanisms. While a turbulent flow regime is characterized by chaotic property changes including rapid variation of flow velocity and pressure, laminar (or streamlined) flow regime In laminar flow, the motion of the particles of the fluid is very orderly as the fluid layers flow in parallel without intermixing. Intuitively, a turbulent regime represents a global exploration of the search space, while a laminar regime indicates local search or exploitation. FRA uses a parameter to balance between its search phases is a direct parallel drawn from Reynold’s number in fluid dynamics (i.e. ratio of inertial to viscous forces acting on a flowing fluid), which is the threshold factor to determine the flow regime of the fluid. Further, the search processes are augmented by the use of arbitrary constants sampled from Levy and Gaussian distributions to increase their randomness.

\subsection{Archimedes' Optimization Algorithm}
\label{AOA}

\begin{figure}[!ht]
    \centering
    \includegraphics[scale=0.9]{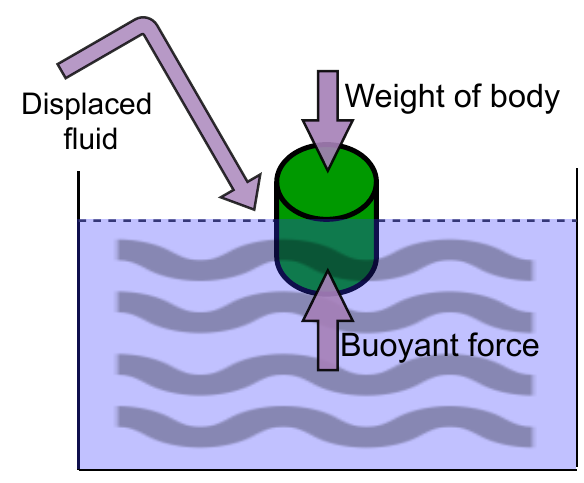}
    \caption{Schematic diagram to represent the Archimedes' principle.}
    \label{fig:arch}
\end{figure}

\noindent Archimedes’ optimization algorithm (AOA) is a recently proposed metaheuristic algorithm by Hashim et al. \cite{hashim2021archimedes} which has its roots in the very famous Archimedes’ Principle of hydrostatics, which states that the upward buoyant force exerted on a body immersed in a fluid (fully or partially) equals to the weight of the fluid displaced by the body. For a body immersed in a fluid, the forces at play are the gravitational and buoyant forces, which are in opposing directions. Accordingly, the body will be at equilibrium if and only if it is acted upon by forces that are balanced i.e. the weight of the body is exactly compensated by the hydrostatic buoyant force. Following Newton’s laws of classical mechanics, the net acceleration of such a body on which balanced forces are acted upon will be zero. This constitutes the optimization paradigm AOA is modelled upon. A schematic diagram showing the scenario is shown in Fig. \ref{fig:arch}. 

\noindent In particular, AOA defines a population to be constituting of candidate agents that represent objects submerged in a fluid of given density. The candidate solutions (i.e. objects) have density, acceleration and volume as their known attributes since these are the quantities on which the buoyant force would depend. The objects are also initialized with random positions which are updated at each step. The acceleration of an object is computed based on the condition of its collision with another object of the population. A higher value of acceleration would imply the search is in the exploration phase (collisions taking place), while a lower value would indicate exploitation (no collision). If a collision takes place, an arbitrary object is taken as a reference for updating the current values, while In case of no collision, the candidate solution follows the global best candidate. AOA uses several parameters to ensure balancing between exploration and exploitation is ably maintained.

\section{Thermodynamics based Metaheuristics}
\label{Thermo}

The branch of thermodynamics deals with the relationship between heat and other forms of energy of the system. There have been several interesting meta-heuristic algorithms developed based on the four laws of thermodynamics. This section focuses on the following algorithms: Thermal exchange optimization, States of matter search, Kinetic gas molecules optimization, and Henry gas solubility optimizer.

\subsection{Thermal Exchange Optimization}
\label{TEO}
Proposed by Kaveh et al. \cite{kaveh2017novel}, thermal exchange optimization (TEO) is modelled on the thermodynamic process of heat transfer. Specifically, it follows Newton’s law of cooling, which states that the rate of heat loss of a body is proportional to the difference in temperatures between the body and its surroundings.

\begin{figure}[ht!]
    \centering
    \includegraphics[width=\linewidth,keepaspectratio]{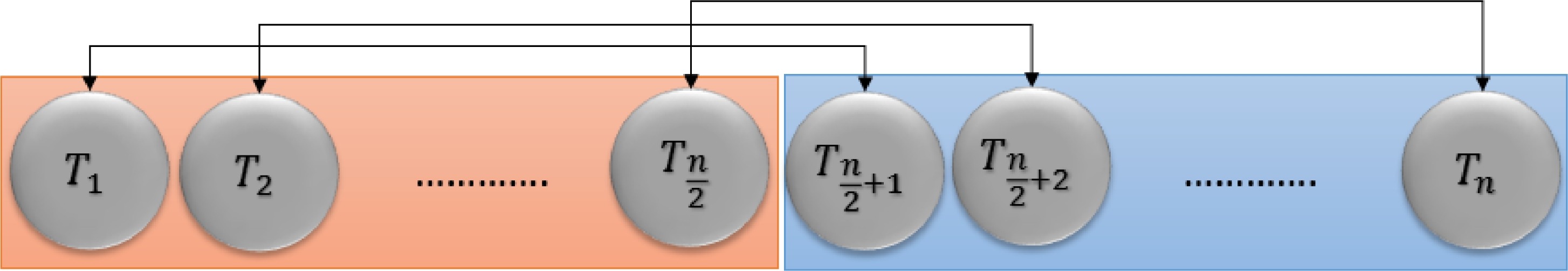}
    \caption{A pictorial description of heat distribution amongst the particles. \textit{Reproduced with permission. Original figure by \cite{kaveh2017novel}}}
    \label{fig:my_label}
\end{figure}

\noindent In TEO, the position of each candidate agent is characterized by its temperature. Each agent is considered as a cooling object that performs thermal energy exchanging with another candidate taken as surrounding fluid. It is worth noting that the choosing of cooling objects and the surrounding environment is similar to the divisive grouping of bodies in CBO (\ref{CBO}). Newton's law of cooling comes into play in order to update the temperatures. Furthermore, TEO also uses two mechanisms for escaping local minima, as well as utilizes memory caching to save a fixed number of recent best solution vectors and their corresponding cost function values, that can be used to enhance the performance of the algorithm without incurring much of a computational cost. 

\subsection{States of Matter Search}
\label{SMS}

The states of matter search (SMS) is a metaheuristic algorithm that bases itself on the simulation of interchangeability among the physical states of matter. Classically speaking, matter exists in three states – gaseous, liquid and solid. The states differ in the extent of their intermolecular forces i.e. solids have the highest intermolecular force of attraction, followed by liquids where the attractive forces are slightly weaker, and then gaseous state where the said forces are very weak in comparison. Accordingly, the packing of the constituent molecules decreases from solids to gases, resulting in variable extents of molecular interactions in the respective physical states. The authors of \cite{cuevas2014optimization} have used this very theoretical backdrop to develop SMS. 

\begin{figure}[!ht]
    \centering
    \includegraphics[width=\linewidth]{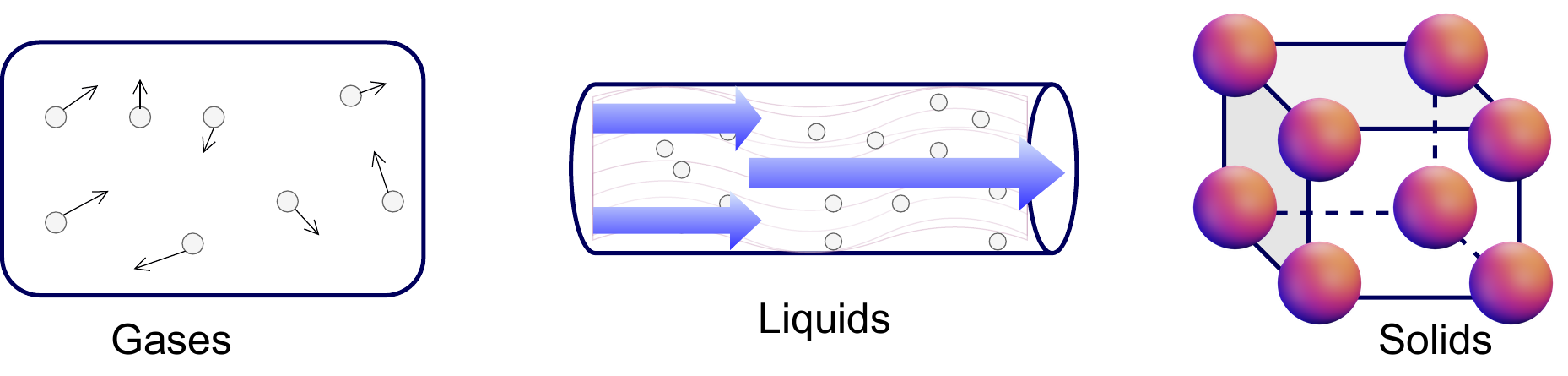}
    \caption{Schematic representation of the packing extents across various states of matter.}
    \label{fig:sms}
\end{figure}

\noindent It is intuitive that gas molecules have greater freedom of randomized movement as compared to liquids and solids. Thus, the gaseous state is modelled as the exploration phase. For solids, since the interactions are very less allowed, they are modelled as the exploitation phase. As for liquids, the molecules exhibit moderate movement and thus represent both exploration and exploitation. SMS follows an evolutionary paradigm where the agents (i.e. molecules) start from the gas state (pure exploration) and the algorithm modifies the exploration and exploitation intensities until the solid state (pure exploitation) is reached. The algorithm incorporates collision and direction vector operators to ensure the balance between exploration and exploitation, while the solution vector updates are performed using a random position operator.

\subsection{Kinetic Gas Molecules Optimization}
\label{KGMO}
Kinetic Gas Molecules Optimization (KGMO) is a swarm based meta-heuristic algorithm based on the kinetic energy of gas molecules. Proposed by Moein et al. \cite{moein2014kgmo}, KGMO uses gas molecules as agents in the search space the performance of each agent is based on their kinetic energy. Each gas molecule in the search space has four specifications: kinetic energy, velocity, position and mass. The velocity and position of the gas molecules are determined using their respective kinetic energies.\\
In the algorithm, after exploration of the entire search space, the gas molecules reach to the point of lowest temperature and kinetic energy because of the attraction between gas molecules originated due to the weak Van der Waal's force between the molecules. KGMO is based on the principle of Boltzmann distribution which implies that the velocity of the gas molecules
is proportional to the exponential of the molecules’ kinetic energy. After the initialisation of each molecule's position and velocity in the search space, the kinetic energy determines the velocity of each gas molecule, which eventually updates the position of each gas molecule.\\
It is noteworthy that both KGMO and Gases Brownian Motion (GBM) optimization algorithm~\cite{abdechiri2013gases} proposed novel ways with regards to evolution of swarm particles. They are completely novel because the evolution of swarm particles was based on newer exploration techniques which were inspired by the non-linear equations which describe thermodynamic systems.

\subsection{Henry Gas Solubility Optimizer}
\label{HGSO}
Proposed by Hashim et al. \cite{hashim2019henry}, the Henry Gas Solubility Optimizer (HGSO) algorithm imitates the huddling behavior of gas to balance exploitation and exploration. The algorithm is modeled on the Henry's law which states that the amount of dissolved gas in a liquid is proportional to its partial pressure above the liquid. Accordingly, the search agents in HGSO are individual gases, each having a partial pressure associated with it.\\
At first the partial pressure of the gases are initialised then a clustering step is applied to cluster the same type of gases. Then from each of the clusters we evaluate the best gases. Calculation of henry's coefficient is followed by calculating the solubility of the gas and subsequently the position is updated. A new type of escape operation is performed in this algorithm is escaping the local optima, in this the worst agents are evaluated and further the position of these worst agents are changed.\\
Also it should be noted that Simulated Annealing and HGSO use the same gas laws. Since HGSO consist of both exploration and exploitation phases, HGSO is generally used for global optimization. Gas coeffient is same for all the gases in same clusters.

\section{Electromagnetism based Metaheuristics}
\label{Electromagnet}

The properties shown by charged particles in electric and magnetic fields provide very interesting insights into global optimization paradigms. Specifically, in this section we discuss some of the optimization techniques based on the principles of classical theories of electrostatics and electromagnetism. They are: Artificial electric field algorithm, Magnetic inspired optimization, Electromagnetic fields algorithm, and Ions motion optimization.

\subsection{Artificial Electric Field Algorithm}
\label{AEFA}
Artificial Electric Field Algorithm (AEFA), proposed by Yadav et al. \cite{yadav2019aefa} is a population-based meta-heuristic algorithm inspired by Coulomb’s laws of electrostatic force. In AEFA, we charge the agents particles and the only force defining the movement of the particles in the search space is the attractive electrostatic force. AEFA proposes an isolated system of particles obeying Coulomb’s law of electrostatic force and laws of motion.\\
After initialising the position of the particles, we assign the charges of the particles using the fitness functions. On the calculation of Coulomb’s constant, and evaluation of the particles based on the fitness values, the force acting on one particle, because of the rest of the particles in the search space, is calculated. The velocity of the particle, which specifies the updated position of each particle in the search space, is based on the acceleration of each particle. The acceleration, caused by the electrostatic force of attraction acting on the particle, is calculated using the second law of motion.\\
For enhancing the exploration capability of the algorithm, the Coulomb’s constant is initialised to a higher value and is decreased with each iteration. AEFA uses both global optima history and local optima history. For updating the position of each particle, in AEFA, the particles are attracted towards the best particle in the search space and the force value specifies the movement of the particle.

\subsection{Magnetic Inspired Optimization}
\label{MIO}

\begin{figure}[ht!]
    \centering
    \includegraphics[scale = 0.8]{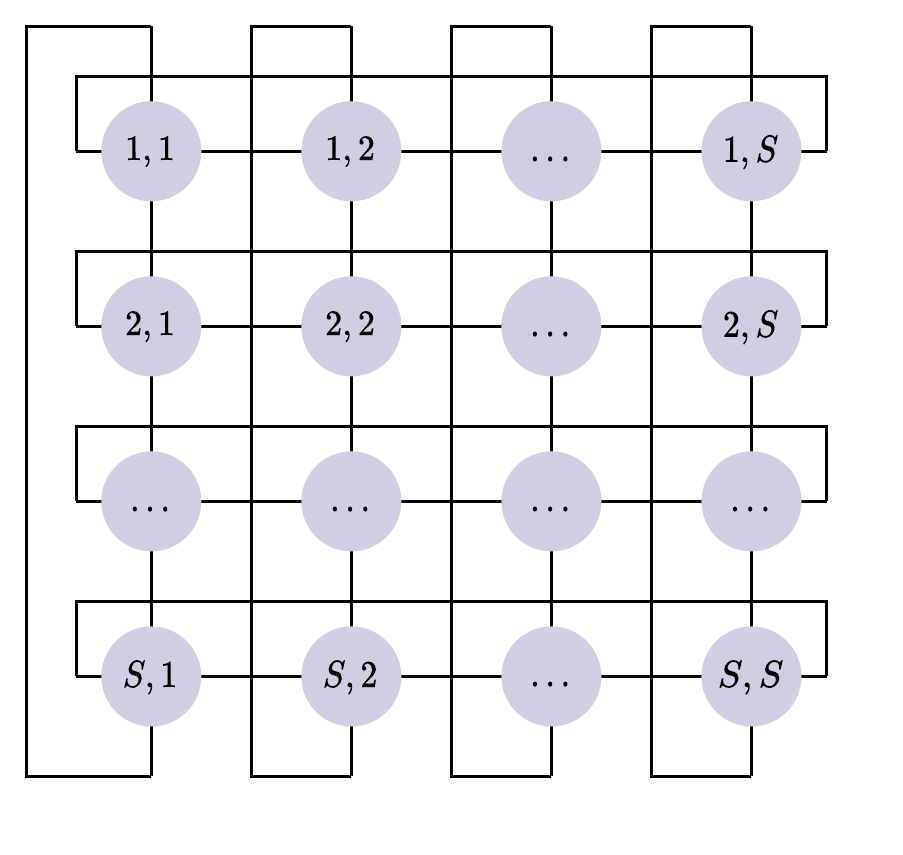}
    \caption{Example of lattice structure in Magnetic Inspired Optimization algorithm. Lines between S charged particles show the connection between neighbour nodes.}
    \label{fig:MIO}
\end{figure}

Magnetic Inspired Optimization (MIO) algorithm is a meta-heuristic algorithm proposed by Tayarani et al. \cite{tayarani2014magnetic} inspired by the magnetic field theory. In the MIO, a magnetic particle scattered in the search space represents each solution. MIO is a cellular-based algorithm with the particles interacting in a lattice-like interactive population as shown in \ref{fig:MIO}. 

\noindent The basic principle behind MIO is the magnetic force acting between the charged particles, which specifies the movement of the particles in the search space. We assign the fitness of each particle in the initialised lattice to the respective magnetic field of the particle. We calculate the mass of each particle based on the normalised magnetic field value. We assign fitter particles with higher mass values, which result in more attractive forces between the neighbouring particles in the lattice. Each particle’s velocity is calculated based on the magnitude of the force and the mass of the particle, which signifies the updated position of the respective particle. After several iterations, the particles around the inferior optimum will start moving towards the superior optimum because of the higher mass of the particles surrounding it.\\
The particles in MIO move towards the particle with higher magnetism (fitness) because of the long-range attractive forces. To overcome the retention of particles around the local optima because of the forces of attraction, MIO also proposes four novel operators: the short-range repulsion, explosion, hybrid explosion-repulsion and crossover interaction, which enhance the performance of the proposed algorithm.

\subsection{Electromagnetic Fields Algorithm}
\label{EFA}
The authors of \cite{abedinpourshotorban2016electromagnetic} presented the Electromagnetic Field Algorithm (EFA), a population-based meta-heuristic approach inspired by the behaviour of electromagnets. An electromagnet, unlike a permanent magnet, produces a magnetic field because of the direction of the electric current and has single polarity (positive/negative). EFA is inspired by the forces of attraction/repulsion acting between electromagnets of different polarities.\\
In EFA, each solution represents an electromagnetic particle formed by the electromagnets. The electromagnets are determined by the variables of the optimization problem. We divide the population into three fields namely positive, negative, and neutral. The attraction-repulsion forces among electromagnets of these three fields specify the movement of the particles towards global minima. A nature-inspired ratio called the golden ratio, which determines the ratio between attractive/repulsive forces, enhances the algorithm. A new electromagnetic particle is formed for every electromagnet (variables) whose fitness is then compared with global minima of the search space, which, if worse, is substituted by the newly formed particle.

\subsection{Ions Motion Optimization}
\label{IMO}
Proposed by Javidy et al. \cite{javidy2015ions}, Ion Motion Optimization (IMO) algorithm was inspired by the nature of the force acting between positively charged (cations) and negatively charged (anions) ions. The algorithm imitates the attractive and repulsive forces between oppositely charged and like-charged ions, respectively. The candidate solution to we group a particular optimization problem into either anions or cations and the nature of the force between the ion specifies their movement in the search space.\\
We evaluated ions based on their fitness values, which are in proportion with the objective function value of the ions. The anions move towards the optimal cation and cations move towards the optimal anion. The amount of movement of the ions depends on the nature of the force between them. The magnitude of the force specifies the momentum of the ions. The requirement of the ions for diversifying and intensifying is met by assuming the ions are in two phases: liquid phase (diversification) and crystal phase(intensification). The exploration capacity of IMO is weaker compared with other approaches like PSO, which are guided by the best solution found so far in the search.

\section{Optics based Metaheuristics}
\label{Optics}
% Optics refers to the way we see in our daily lives, many of us require assistance of artificial lenses to correct our vision. Hence we see the world in a much better manner. With inspirations of these, researchers have proposed metaheurtics based on optics. -- ETA KI CHILO ?!?!?!?!? SHURUTEI ART. LENS ??

Optics deals with the behaviour and properties of light and its interaction with matter. The algorithms discussed in this section are based on the properties of the systems in optics. The two approaches are: Ray optimization and Optics inspired optimization.
\subsection{Ray Optimization}
\label{RO}
This is the first work in the world of metaheuristics derived from ray optics, proposed by Kaveh et al. \cite{kaveh2012new}. The algorithm is inspired by Snell’s law of refraction of light, which explains why and how the direction of a light ray changes when it faces a change in medium i.e. while passing from one medium to another. Ray Optimization (RO) specifically uses ray tracing as the backbone of its search mechanism using basic geometrical operations and is formulated in three steps: (1) scattering; (2) ray movement and motion refinement; and (3) ray convergence.\\
The initial population pool of RO comprises rays of light represented as candidate vectors. A candidate solution in RO follows the path traced by a ray to move about in the search space. Since ray-tracing is feasible only in two and three-dimensional spaces, for higher dimensions the solution vector is divided into groups comprising two or three members, which are then moved to new positions following the exploration and exploitation procedures described by the algorithm. Thus, each update of the candidate solution consists of position updates of its constituent groups in 2D or 3D spaces. The best positions of the local vectors along with the best global solution vector are stored in each iteration until the termination condition is reached.

\subsection{Optics Inspired Optimization}
\label{OIO}
\begin{figure}[ht!]
    \centering
    \includegraphics[width=\linewidth,keepaspectratio]{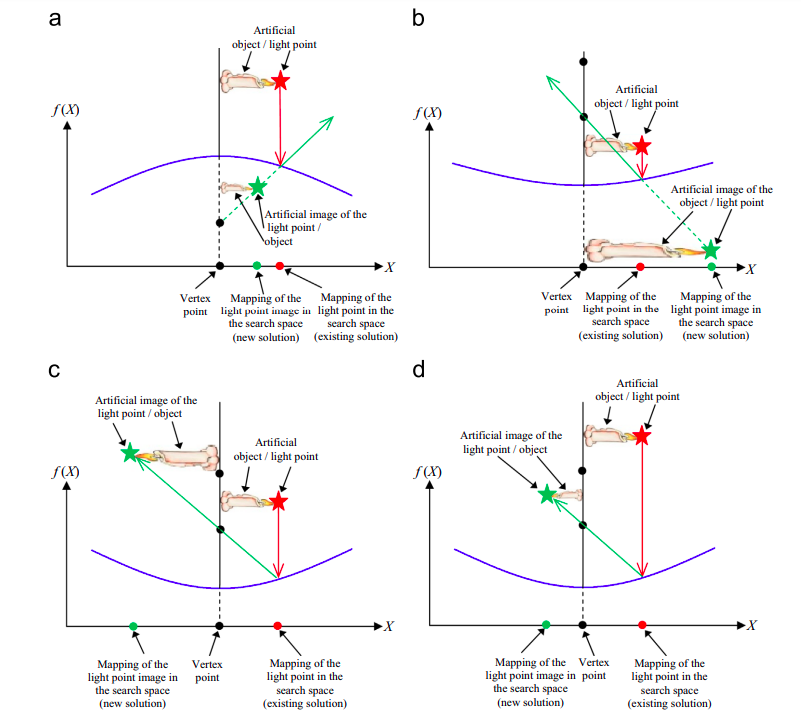}
    \caption{The processes involved in OIO.(a) This is when the function serves as a convex surface (b) The function as concave surface and the artificial object/light point is between artificial focal point and the function surface. (c) Function as concave surface and the artificial object/light point is between artificial focus and centre of curvature. (d) The function as as a concave surface and the artificial object/light point is beyond centre of curvature. \textit{Reproduced with permission. Original figure by \cite{kashan2015new}}}
    \label{fig:OIO}
\end{figure}

Optics-inspired Optimization (OIO) was proposed by Kashan \cite{kashan2015new} that has been developed based on reflection of light by spherical mirrors. These mirrors have a very interesting property to converge and diverge images of the specific objects. In OIO the search field of the interested problem is to be optimized as a wavy mirror in which the concave mirror is represented as a valley and the convex mirror is represented as a peak. In other words, the surface of the objective function is considered a spherical mirror that can either be converging (i.e. concave) or diverging (i.e. convex), and each candidate solution represents an artificial light point, from which the new solutions are generated based on the final produced image upon reflection. The detailed description of the approach can be found in Fig \ref{fig:OIO}.

\section{Other Physics-based Metaheuristic Algorithms}
\label{Misc.}
% There are quite of lot of state-of-the-art optmization algorithms which are inspired by several physical phenomenon but cannot be grouped into the categories. Some of them are discussed underneath
In this section, we extend our discussion to some of the more unique physics-inspired optimization techniques that which are too diverse to be categorised under the domains explored in this chapter. The discussed algorithms are: Simulated annealing, Multi-verse optimizer, Atom search optimization, and Nuclear reaction optimization.

\subsection{Simulated Annealing}
\label{SA}
Simulated Annealing (SA), proposed by Kirkpatrick et al. \cite{kirkpatrick1983optimization}, was the first meta-heuristic algorithm based on non-linear physics processes. Proposed in the early 1980s, this algorithm was inspired by the annealing process in material science which deals with obtaining the configurations of the material with minimum energy of molecules or atoms of the material. SA imitates the annealing process in a way that the molecules of the material represent the candidate solution of the optimization problem and the energy of the system is used to signify the fitness function.\\
The basic principle of SA is to obtain the optimal state through lowering the temperature of the system which is used as a control parameter for the optimization problem. The primary approach is to consider the highest probability state of a system which would be in thermal equilibrium, which would have an energy $\epsilon$, proportional to the Boltzmann probability factor($e^{\frac{-\epsilon_i}{kT}}$).\\
On modification of each state and its corresponding neighbours, if a decreasing change is observed in the evaluation of the fitnesses of the states, then the change is accepted. On decreasing the temperature of the system with each iteration, the probability of acceptance of worse solution decreases.\\
It is to be noted that, while majority of meta-heuristic algorithms, which have been proposed, have focused on the enhancement of the exploration phase of the searching problem, SA focuses on enhancing the exploitation capability through the implementation of non-linear physics process.

\subsection{Multi-verse Optimizer}
\label{MVO}
Inspired by the theory of alternate universes in cosmology, Multi-verse Optimizer (MVO) is a meta-heuristic algorithm proposed by Mirjalili et al. \cite{mirjalili2016multi}. The multiverse theory suggests the existence of other universes besides the one we are living in, because of multiple big bangs in the beginning. Each of these universes has an inflation rate which defines their expansion rate in space, which results in the interaction and even collision between these universes. The algorithm is primarily based on three concepts of multi-verse theory, namely black-hole, white hole and wormhole.\\
We consider each population in MVO as a universe where each variable in the universe represents a cosmological object in that universe. We assign each candidate solution an inflation rate, proportional to the corresponding fitness value of the particle. Universes may have a movement of particles through black/white tunnels between them. We consider a universe having more white holes if it has a higher inflation rate, and black holes if it has a lower inflation rate. Objects move towards lower inflation rates from higher inflation rates, which enhances the average inflation rate.\\
After sorting the universes according to their inflation rates, the assignment of a white hole to a particular universe has been done through a roulette wheel mechanism, based on the normalised inflation rates. To enhance exploitation performance, a wormhole tunnel enables the exchange of objects between a universe and the best universe formed so far.
 
\subsection{Atom Search Optimization}
\label{ASO}
\begin{figure}[ht!]
    \centering
    \includegraphics[keepaspectratio]{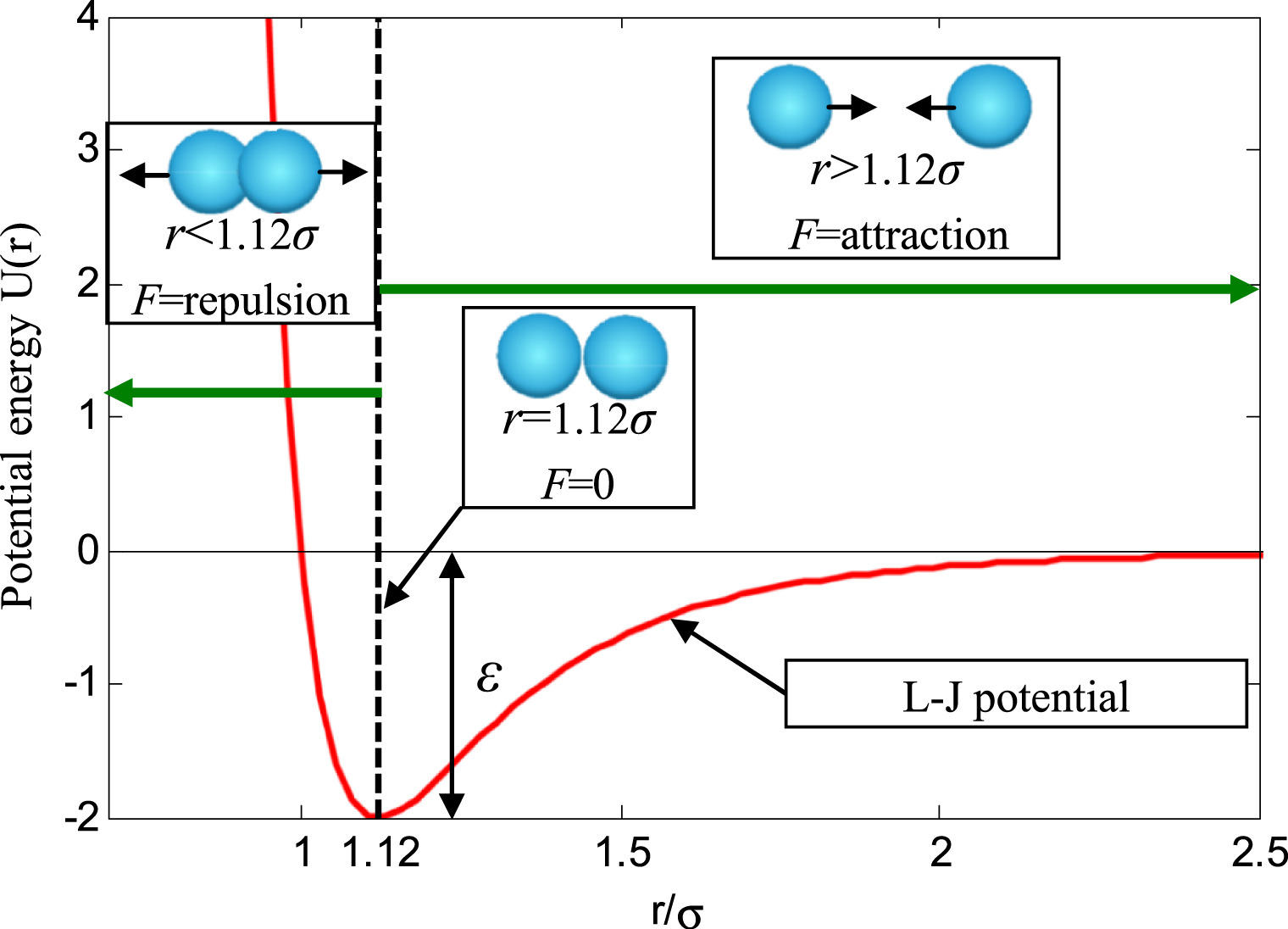}
    \caption{A pictorial illustration to how ASO is modelled upon. \textit{Reproduced with permission. Original figure by \cite{zhao2019atom}}}
    \label{fig:ASO}
\end{figure}

The Atom Search Optimization (ASO), proposed by Zhao et al. \cite{zhao2019atom} is a meta-heuristic optimization algorithm inspired by basic molecular dynamics. ASO is a population-based meta-heuristic algorithm that mimics the atomic motion determined by interaction and constraint forces.\\
In ASO, we represent each solution by the position of the atom in the search space, whose value is determined by the mass of the atom. The heavier atoms seek to attract the other atoms of the search space towards them. The lower value of acceleration in heavier atoms ensures a better search for the local optima, while the higher value of acceleration in the case of lighter atoms enables the search for better promising regions in the entire search space. The acceleration of atoms comes from two parts, (i) interaction force caused by the L-J potential, and (ii) the constraint force caused by bond-length potential, which is the weighted position difference between each atom and the best atom. A pictorial illustration of this can be found in Fig. \ref{fig:ASO}.

\noindent In the initial stages of the algorithm, the atoms interact with each other through attractive and repulsive forces depending on the distance between them, which results in the lighter atoms being attracted towards the heavier atoms. The repulsive force denies the over-concentration of atoms around a local optimum, ensuring the exploration capability of the algorithm throughout the search space. With each iteration, the repulsive force grows weaker and the attractive forces grow stronger, ensuring the exploitation capabilities of the algorithm. With each iteration, the velocities and position of each particle are updated until the stopping criterion is met.

\subsection{Nuclear Reaction Optimization}
\label{NRO}
Nuclear Reaction Optimization (NRO) is a meta-heuristic algorithm based on the phenomena of nuclear reaction and was proposed by Wei et al. \cite{wei2019nuclear}. The NRO comprises two phases: the Nuclear Fusion (NFu) phase and the Nuclear Fission (NFi) phase. NFi inhibits the exploitation characteristics to find a better solution near the local optima while NFu exhibits exploration capabilities for obtaining a global solution.\\
The NRO algorithm assumes that the whole search space is in a sealed container where all nuclei interact with each other with the nuclei representing the variables of the optimization problem. In the NFi phase, NRO adopts the Gaussian walk method to generate new nuclei after fission. The gaussian walk is primarily implemented to imitate the fission fragments with different states for enhanced exploitation of a search space after which we update the position of the nuclei using the fitness function.\\
We divide the NFu phase into two phases ionization and fusion. The nuclei are evaluated based on their fitnesses and the fitter nuclei are retained for the enhanced exploitation of the NFu phase, while the less fitter nuclei are used to improve exploration. We implement the fusion stage movement to change the state of an ion combined with information of other ions in the nuclear reaction population. Similar to the ionization phase, this phase uses the variants of difference operators for enhanced exploration. In the fusion phase, Levy flight is implemented to assist the current solution in escaping from the local optimal space.

\section{Conclusion}
\label{Conc.}
In this chapter we have discussed the most recent and widely-used metaheuristic algorithms based on non-linear physics. The algorithms have been categorized according to the branch of physics they have drawn inspiration from like classical mechanics, thermodynamics, electromagnetism among others. We have described the intuition behind each of the algorithms categorised under 6 domains that include both recent as well as widely-used proposals, emphasizing on how each algorithm imitates the non-linear physical processes for finding the optimal solution. We hope this chapter would provide a foundation of physics-inspired optimization paradigms for the readers which would encourage them to explore further into the vast world of metaheuristic algorithms.

\bibliographystyle{splncs03}
\bibliography{references}

\begin{thebibliography}{10}
\providecommand{\url}[1]{\texttt{#1}}
\providecommand{\urlprefix}{URL }

\bibitem{abdechiri2013gases}
Abdechiri, M., Meybodi, M.R., Bahrami, H.: Gases brownian motion optimization:
  an algorithm for optimization (gbmo). Applied Soft Computing  13(5),
  2932--2946 (2013)

\bibitem{abedinpourshotorban2016electromagnetic}
Abedinpourshotorban, H., Shamsuddin, S.M., Beheshti, Z., Jawawi, D.N.:
  Electromagnetic field optimization: a physics-inspired metaheuristic
  optimization algorithm. Swarm and Evolutionary Computation  26,  8--22 (2016)

\bibitem{yadav2019aefa}
Anita, Yadav, A.: Aefa: Artificial electric field algorithm for global
  optimization. Swarm and Evolutionary Computation  48,  93--108 (2019)

\bibitem{chakraborti2015statistical}
Chakraborti, A., Challet, D., Chatterjee, A., Marsili, M., Zhang, Y.C.,
  Chakrabarti, B.K.: Statistical mechanics of competitive resource allocation
  using agent-based models. Physics Reports  552,  1--25 (2015)

\bibitem{cuevas2014optimization}
Cuevas, E., Echavarr{\'\i}a, A., Ram{\'\i}rez-Orteg{\'o}n, M.A.: An
  optimization algorithm inspired by the states of matter that improves the
  balance between exploration and exploitation. Applied intelligence  40(2),
  256--272 (2014)

\bibitem{demetrius2013boltzmann}
Demetrius, L.A.: Boltzmann, darwin and directionality theory. Physics reports
  530(1),  1--85 (2013)

\bibitem{dougan2015new}
Do{\u{g}}an, B., {\"O}lmez, T.: A new metaheuristic for numerical function
  optimization: Vortex search algorithm. Information Sciences  293,  125--145
  (2015)

\bibitem{faramarzi2020equilibrium}
Faramarzi, A., Heidarinejad, M., Stephens, B., Mirjalili, S.: Equilibrium
  optimizer: A novel optimization algorithm. Knowledge-Based Systems  191,
  105190 (2020)

\bibitem{formato2007central}
Formato, R.A.: Central force optimization. Prog Electromagn Res  77(1),
  425--491 (2007)

\bibitem{hashim2019henry}
Hashim, F.A., Houssein, E.H., Mabrouk, M.S., Al-Atabany, W., Mirjalili, S.:
  Henry gas solubility optimization: A novel physics-based algorithm. Future
  Generation Computer Systems  101,  646--667 (2019)

\bibitem{hashim2021archimedes}
Hashim, F.A., Hussain, K., Houssein, E.H., Mabrouk, M.S., Al-Atabany, W.:
  Archimedes optimization algorithm: a new metaheuristic algorithm for solving
  optimization problems. Applied Intelligence  51(3),  1531--1551 (2021)

\bibitem{javidy2015ions}
Javidy, B., Hatamlou, A., Mirjalili, S.: Ions motion algorithm for solving
  optimization problems. Applied Soft Computing  32,  72--79 (2015)

\bibitem{kashan2015new}
Kashan, A.H.: A new metaheuristic for optimization: optics inspired
  optimization (oio). Computers \& Operations Research  55,  99--125 (2015)

\bibitem{kaveh2017novel}
Kaveh, A., Dadras, A.: A novel meta-heuristic optimization algorithm: thermal
  exchange optimization. Advances in Engineering Software  110,  69--84 (2017)

\bibitem{kaveh2012new}
Kaveh, A., Khayatazad, M.: A new meta-heuristic method: ray optimization.
  Computers \& structures  112,  283--294 (2012)

\bibitem{kaveh2014colliding}
Kaveh, A., Mahdavi, V.R.: Colliding bodies optimization: a novel meta-heuristic
  method. Computers \& Structures  139,  18--27 (2014)

\bibitem{kennedy1995particle}
Kennedy, J., Eberhart, R.: Particle swarm optimization. In: Proceedings of
  ICNN'95-international conference on neural networks. vol.~4, pp. 1942--1948.
  IEEE (1995)

\bibitem{kirkpatrick1983optimization}
Kirkpatrick, S., Gelatt, C.D., Vecchi, M.P.: Optimization by simulated
  annealing. science  220(4598),  671--680 (1983)

\bibitem{mirjalili2016multi}
Mirjalili, S., Mirjalili, S.M., Hatamlou, A.: Multi-verse optimizer: a
  nature-inspired algorithm for global optimization. Neural Computing and
  Applications  27(2),  495--513 (2016)

\bibitem{mirjalili2014grey}
Mirjalili, S., Mirjalili, S.M., Lewis, A.: Grey wolf optimizer. Advances in
  engineering software  69,  46--61 (2014)

\bibitem{moein2014kgmo}
Moein, S., Logeswaran, R.: Kgmo: A swarm optimization algorithm based on the
  kinetic energy of gas molecules. Information Sciences  275,  127--144 (2014)

\bibitem{rashedi2009gsa}
Rashedi, E., Nezamabadi-Pour, H., Saryazdi, S.: Gsa: a gravitational search
  algorithm. Information sciences  179(13),  2232--2248 (2009)

\bibitem{tahani2019flow}
Tahani, M., Babayan, N.: Flow regime algorithm (fra): a physics-based
  meta-heuristics algorithm. Knowledge and Information Systems  60(2),
  1001--1038 (2019)

\bibitem{tayarani2014magnetic}
Tayarani-N, M.H., Akbarzadeh-T, M.R.: Magnetic-inspired optimization
  algorithms: Operators and structures. Swarm and Evolutionary Computation  19,
   82--101 (2014)

\bibitem{wei2019nuclear}
Wei, Z., Huang, C., Wang, X., Han, T., Li, Y.: Nuclear reaction optimization: A
  novel and powerful physics-based algorithm for global optimization. IEEE
  Access  7,  66084--66109 (2019)

\bibitem{zhao2019atom}
Zhao, W., Wang, L., Zhang, Z.: Atom search optimization and its application to
  solve a hydrogeologic parameter estimation problem. Knowledge-Based Systems
  163,  283--304 (2019)

\end{thebibliography}

\end{document}